%% file: emnlp_camera_ready.tex
\pdfoutput=1
\documentclass[11pt]{article}
\usepackage[]{acl}
\usepackage{times}
\usepackage{latexsym}
\usepackage[T1]{fontenc}
\usepackage[utf8]{inputenc}
\usepackage{microtype}
\usepackage{inconsolata}
\usepackage{graphicx}
\usepackage{hyperref}       % hyperlinks
\usepackage{url}            % simple URL typesetting
\usepackage{booktabs}       % professional-quality tables
\usepackage{amsfonts}       % blackboard math symbols
\usepackage{nicefrac}       % compact symbols for 1/2, etc.
\usepackage{microtype}      % microtypography
\usepackage{xcolor}         % colors
\usepackage{float}
\usepackage{amsmath}
\usepackage{amssymb}
\usepackage{mathtools}
\usepackage{amsthm}
\usepackage{graphicx}
\usepackage{wrapfig, lipsum, booktabs}
\usepackage{color, colortbl}
\usepackage{bbm}
\usepackage{subcaption}
\usepackage{multirow}
\usepackage{tikz}
\usepackage{enumitem}
\usepackage[ruled,vlined]{algorithm2e}
\usepackage{algorithmic}
\definecolor{citecol}{HTML}{00008B}
\definecolor{tableofcontent}{HTML}{E63E15}
\definecolor{urlcol}{HTML}{2470D8}
\usepackage{hyperref}
\usepackage{multirow}
\usepackage[normalem]{ulem}
\useunder{\uline}{\ul}{}
\usepackage{nicematrix}
\theoremstyle{plain}

\theoremstyle{definition}

\theoremstyle{remark}
\usepackage{arydshln}

\usepackage{hyperref}
\usepackage{mdframed}
\usepackage{url}
\hypersetup{
    colorlinks=true,       % false: boxed links; true: colored links
    linkcolor=tableofcontent, 
    citecolor=citecol,        % color of links to bibliography
    urlcolor=black,           % color of external links
}
\definecolor{Gray}{gray}{0.9}
\definecolor{Gray}{gray}{0.9}
\newcommand{\xhdr}[1]{{\vspace{1pt}\noindent\bfseries #1}.}
\newcommand{\ie}{\textit{i.e., }}
\newcommand{\eg}{\textit{e.g., }}

\title{GRIL: Knowledge Graph Retrieval-Integrated Learning \\
with Large Language Models}

\author{Jialin Chen$^{1}$, Houyu Zhang$^{2}$, Seongjun Yun$^{2}$, Alejandro Mottini$^{2}$, Rex Ying$^{1}$,\\ 
\textbf{Xiang Song$^{2}$, Vassilis N. Ioannidis$^{2}$, Zheng Li$^{2}$, Qingjun Cui$^{2}$}\\
Yale University$^1$\qquad Amazon$^{2}$}

\begin{document}
\maketitle
\input{chapters/00abstract}
\input{chapters/01introduction}
\input{chapters/02relatedwork}

\input{chapters/03method}

\input{chapters/04experiments}
\input{chapters/05conclusion}

\bibliography{anthology}

\appendix
\clearpage
\input{chapters/06appendix}
\end{document}

%% file: chapters/00abstract.tex
\begin{abstract}
    Retrieval-Augmented Generation (RAG) has significantly mitigated the hallucinations of Large Language Models (LLMs) by grounding the generation with external knowledge. Recent extensions of RAG to graph-based retrieval offer a promising direction, leveraging the structural knowledge for multi-hop reasoning.
    However, existing graph RAG typically decouples retrieval and reasoning processes, which prevents the retriever from adapting to the reasoning needs of the LLM. They also struggle with scalability when performing multi-hop expansion over large-scale graphs, or depend heavily on annotated ground-truth entities, which are often unavailable in open-domain settings.
    To address these challenges, we propose a novel graph retriever trained end-to-end with LLM, which features an attention-based growing and pruning mechanism, adaptively navigating multi-hop relevant entities while filtering out noise. 
    Within the extracted subgraph, structural knowledge and semantic features are encoded via soft tokens and the verbalized graph, respectively, which are infused into the LLM together, thereby enhancing its reasoning capability and facilitating interactive joint training of the graph retriever and the LLM reasoner. Experimental results across three QA benchmarks show that our approach consistently achieves state-of-the-art performance, validating the strength of joint graph–LLM optimization for complex reasoning tasks. Notably, our framework eliminates the need for predefined ground-truth entities by directly optimizing the retriever using LLM logits as implicit feedback, making it especially effective in open-domain settings.
\end{abstract}

%% file: chapters/01introduction.tex
\section{Introduction}
Large Language Models (LLMs) have shown remarkable abilities in natural language processing tasks~\cite{brown2020language, llama3, achiam2023gpt}. Despite their success, LLMs often suffer from hallucinations, generating outputs that may be factually incorrect, particularly in scenarios requiring domain-intensive knowledge. To mitigate hallucinations and improve domain-specific performance, recent approaches have explored the Retrieval-Augmented Generation (RAG) framework, which enhances LLMs by retrieving external knowledge~\cite{gao2023retrieval, lewis2020retrieval,wu2023retrieve,fan2024survey} and has proven especially beneficial in tasks such as Knowledge Graph Question Answering (KGQA)~\cite{bao2016constraint, huang2019knowledge,zheng2017natural}.

% Unlike traditional text-based retrieval, KGQA requires the retriever to navigate complex, multi-hop relationships embedded within a knowledge graph (KG) while ensuring the retrieved information is both relevant and useful. Consequently, KGQA tasks demand not only an accurate structure-aware retriever but also deep reasoning capabilities to correctly interpret the extracted knowledge~\cite{sun2019pullnet, li2023graph,pan2024unifying,yani2021challenges,yasunaga2021qa}.
% This shift to graph-based RAG is motivated by the fact that knowledge graphs offer cleaner, less ambiguous relational information than unstructured text, making them more efficient for reducing hallucinations.

Unlike traditional knowledge bases such as documents and textbooks, knowledge graphs (KGs) provide cleaner and well-structured relational knowledge, offering a more precise and efficient base for navigating complex reasoning paths and reducing hallucinations compared to unstructured data. Consequently, recent efforts have extended RAG by incorporating graph retrieval, where knowledge graphs—structured and domain-specific knowledge bases—are used to guide retrieval and reasoning. Existing approaches use LLMs as a retriever for KGQA tasks (abbreviated as \textit{LLM-as-Retriever})~\cite{ROG,TOG,KDCOT}, extracting relevant facts or relation paths from the KGs based on LLMs' internal knowledge. However, such \textit{LLM-as-retriever} approaches are typically ill-equipped to fully exploit the intricate structures within KGs, such as multi-hop logical relations, and require multiple calls to process different parts of the graph, which introduces scalability issues when dealing with large-scale KGs. 
There have also been attempts using a graph neural network (GNN) during retrieval (abbreviated as \textit{GNN-as-Retriever})~\cite{he2024g, hu2024grag, mavromatis2024gnnrag}. However, one limitation is that most existing approaches train the retrieval and reasoning components separately, resulting in a disjoint optimization process where the retriever focuses solely on relevant facts without aligning with the reasoning needs of the LLM. Moreover, many rely on ground truth retrieval paths for retriever training, limiting their applicability to open-domain scenarios where the ground truth entities are not available. 
% Additionally, while some approaches have attempted to incorporate structural information during retrieval, they often fail to capture the full depth of the graph's relationships, limiting the LLM’s ability to reason over the graph’s intricate structure.

\begin{figure}[ht]
    \centering
    \includegraphics[width=\linewidth]{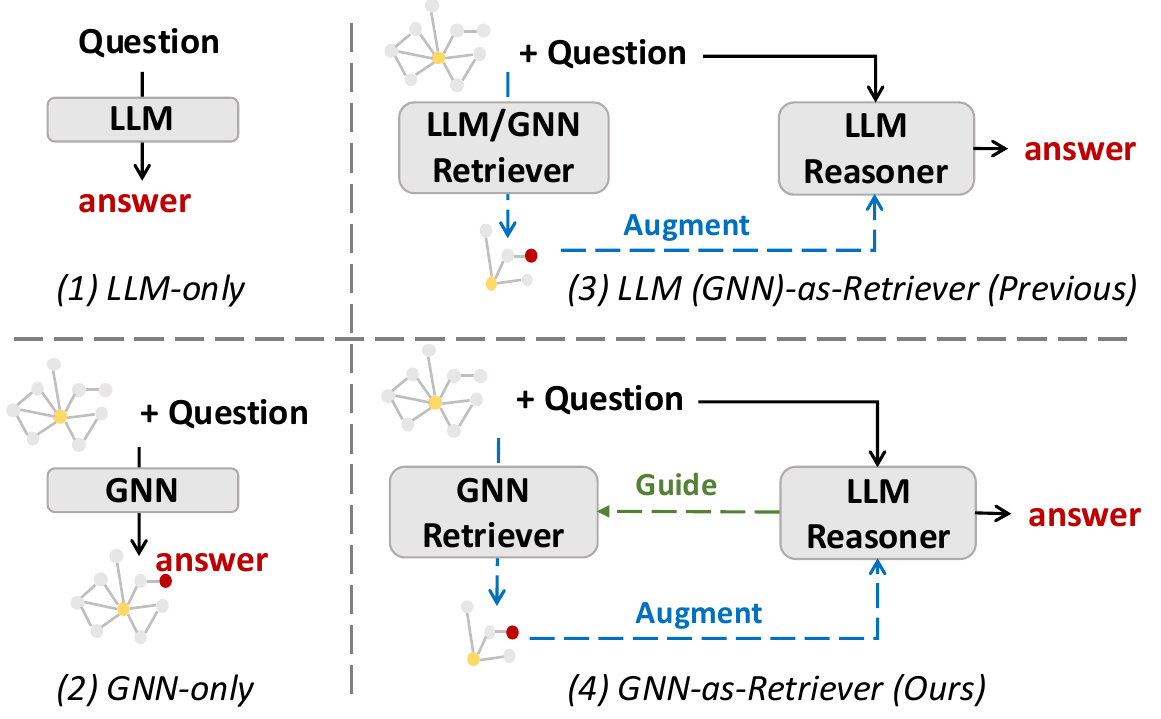}
    \caption{The landscape of existing methods. \textit{(1) LLM-only} and \textit{(2) GNN-only} approaches use a single LLM or GNN to predict the answer. \textit{(3) LLM (GNN)-as-Retriever} approaches rely on the RAG framework to reduce the hallucination and improve the accuracy of LLMs' output. Different from previous approaches, we utilize the LLM reasoner to supervise the GNN retriever, improving the retrieval quality and reasoning accuracy.}
    \label{fig:landscape}
\end{figure}
To address these limitations, we propose \textbf{GRIL}, a novel framework that enables \textbf{G}raph \textbf{R}etrieval-\textbf{I}ntegrated \textbf{L}earning with LLM in an end-to-end manner. As illustrated in Figure~\ref{fig:landscape}, our approach extends beyond conventional retriever-augmented reasoning by introducing a reverse feedback loop from the LLM to guide the retriever. This LLM feedback is essential for the model's generalizability in open-domain scenarios, by providing a complementary supervision signal, thus eliminating its reliance on answer entities during retriever training. Intuitively, the reverse feedback shifts retrieval from mere relevance to actual usefulness for LLM reasoning. Moreover, we introduce a novel graph retriever that iteratively grows and prunes the extracted knowledge subgraph, which allows the retrieval process to focus on the most relevant multi-hop entities while filtering out irrelevant information, improving retrieval efficiency and accuracy. The proposed integrated training ensures the graph retriever and LLM reasoner are tightly coupled and jointly optimized, fostering better synergy between retrieval and reasoning and improving overall performance.

% This framework leverages the power of graph attention mechanisms to capture the \jialin{intricate structural relationships => be more concrete }within knowledge graphs, allowing the retrieval process to dynamically focus on the most relevant multi-hop entities while filtering out irrelevant information. This attention-based retrieval improves the quality of the extracted knowledge subgraph provided to the LLM, enabling it to reason more effectively over the graph's topology. Additionally, our approach enhances generalizability by incorporating implicit feedback from the LLM itself to supervise the retriever, instead of relying on ground truth entities for training. This adaptive feedback loop enables the model to handle open-domain KGQA tasks, and ensures that the retriever can dynamically adjust to a wider range of queries and datasets. Furthermore, the framework supports end-to-end joint training of both the graph retriever and the LLM reasoner, ensuring that these components are tightly coupled and optimized together. This integrated training process enhances the synergy between retrieval and reasoning, enabling the system to retrieve more relevant knowledge for the reasoning needs of the LLM  and improving overall performance on knowledge-intensive tasks.

Our contributions are threefold. (1) We introduce a novel framework that combines an adaptive attention-based graph retriever with joint training alongside the LLM reasoner, improving the accuracy and relevance of knowledge retrieval. (2) Experiments demonstrate performance improvement over competitive models, achieving state-of-the-art results on KGQA benchmarks, and showing strong open-domain generalizability, where most baselines fail. (3) We address the inefficiency of multiple LLM calls during the retrieval process and enable small BERT-level language models to match or even outperform 7B LLMs when paired with our graph retriever, improving inference efficiency and reducing deployment costs significantly.

%% file: chapters/02relatedwork.tex
\section{Related Work}
\textbf{Knowledge Graph Question Answering} aims to answer natural language queries based on a structured knowledge graph (KG), which consists of entities and their relationships.~\cite{sun2019pullnet, li2023graph,pan2024unifying,yani2021challenges, yasunaga2021qa,reinanda2020knowledge}. The main challenge in KGQA is handling complex reasoning and mapping it accurately to relevant subgraphs in the KG. Traditional approaches to KGQA often utilize Graph Neural Networks (GNNs) to learn embeddings for entities by aggregating information from their neighbors, supervised by labels that indicate whether a node is an answer for a given question~\cite{yasunaga2021qa,mavromatis2022rearev, NSM}. These methods typically lack the ability to effectively traverse long, complex paths and multi-hop reasoning, leading to limited expressivity in capturing deeper structural information. Moreover, they cannot generalize well to open-domain settings where the ground-truth answers are not covered by the KGs. Our method addresses this issue through a dynamic, attention-based graph retriever that iteratively grows and prunes the subgraph, enabling better exploration of multi-hop relationships while eliminating the dependency on ground truth entities during training of the graph retriever.

\xhdr{Retrieval-augmented LLM Reasoning}
To address the hallucination issue within the LLMs' output, recent works have explored a retrieval-augmented generation (RAG) framework, where a retriever extracts relevant information from an external knowledge base (\eg text corpus, a KG, or other structured resources) and converts it into textual prompts for LLMs~\cite{gao2023retrieval,li2024matching}. Compared with traditional RAG systems applied to document-based knowledge bases~\cite{robertson2009probabilistic, karpukhin2020dense}, RAG on knowledge graphs (KGs) provide cleaner and more well-structured relations with less ambiguity for LLM reasoning. Some works focus on leveraging LLMs as retrievers to extract relevant facts or relational paths from a graph~\cite{wu2023retrieve, TOG, ROG, yu2022decaf}, which are then used for reasoning. However, these approaches often require multiple LLM calls, making them computationally expensive and inefficient. Recent works also attempt to amalgamate graph retrievers and LLMs~\cite{peng2024graph,he2024g, hu2024grag, mavromatis2024gnnrag}, incorporating structural information during the retrieval process to improve performance. However, the graph retriever and LLMs are typically optimized separately, which limits their ability to fully exploit the synergy between the graph structure and LLMs' reasoning capabilities. Our proposed GRIL addresses the above issues by enabling end-to-end training between the GNN retriever and the LLM reasoner, optimizing both components in an interactive way.

% Logic: Why RAG (General KB) -> Why Graph RAG -> Previous methods (LLM-Retriever -> GNN-Retriever)

% Previous: 
% As KGs usually contain millions of edges and nodes that exceed the context window size of LLM, recent efforts instead utilize a retriever that retrieves relevant information from the KG and converts it into textual prompts for LLM augmented reasoning, in a retrieval-augmented generation (RAG) style~\cite{wu2023retrieve, TOG, ROG, yu2022decaf}. There have been works that request an LLM to explore the KG and retrieve relevant facts or relation paths. The LLM is typically further finetuned to generate answers based on the retrieved information and the given question. However, these methods require multiple LLM calls during the RAG process. Moreover, LLMs are typically ill-equipped to fully exploit the intricate structural relationships inherent in the KG, leading to suboptimal performance in tasks that require deep structural understanding. Recent works attempt to amalgamate graph retrievers and LLMs~\cite{he2024g, hu2024grag, mavromatis2024gnnrag}, incorporating structural information during the retrieval process to improve performance. However, the graph retriever and LLM are typically fine-tuned separately, which limits their ability to fully exploit the synergy between the graph’s structure and the language model's reasoning capabilities.

%% file: chapters/03method.tex
\section{Preliminaries and Background}
Knowledge Graph Question Answering (KGQA) approaches aim to predict the correct answer $a$, given a question $q$ and a KG $\mathcal{G}$ that provides relevant reasoning information. To reduce hallucination, a graph retriever extracts a subgraph $\mathcal{G}_s\subseteq\mathcal{G}$ that is the most relevant and useful for answering the question. The target is to learn a model that optimizes the conditional probability:
\begin{equation}\label{eq:prob_rag}
   \small p(a|\mathcal{G},q) = \sum_{\mathcal{G}_s}p_{\phi}(a|\mathcal{G}_s, q)p_{\theta}(\mathcal{G}_s|q, G),
\end{equation}
where $p_{\theta}$ estimates the prior distribution on an extracted subgraph $\mathcal{G}_s$ conditioned on the given query $q$, and $p_\phi$ indicates the likelihood of the answer $a$ given the query $q$ and the subgraph $\mathcal{G}_s$, predicted by a reasoner (\eg LLMs). Maximizing the log-likelihood decouples the retriever from the reasoner as follows,
\begin{equation}\label{eq:log_prob}
\small \mathcal{L}=\max_{\phi, \theta}\sum_{(q,a,\mathcal{G}_s)}\log p_{\phi}(a|\mathcal{G}_s, q)+\log p_{\theta}(\mathcal{G}_s|q).
\end{equation}
\section{Methodology}
\begin{figure}[ht]
    \centering
    \includegraphics[width=\linewidth]{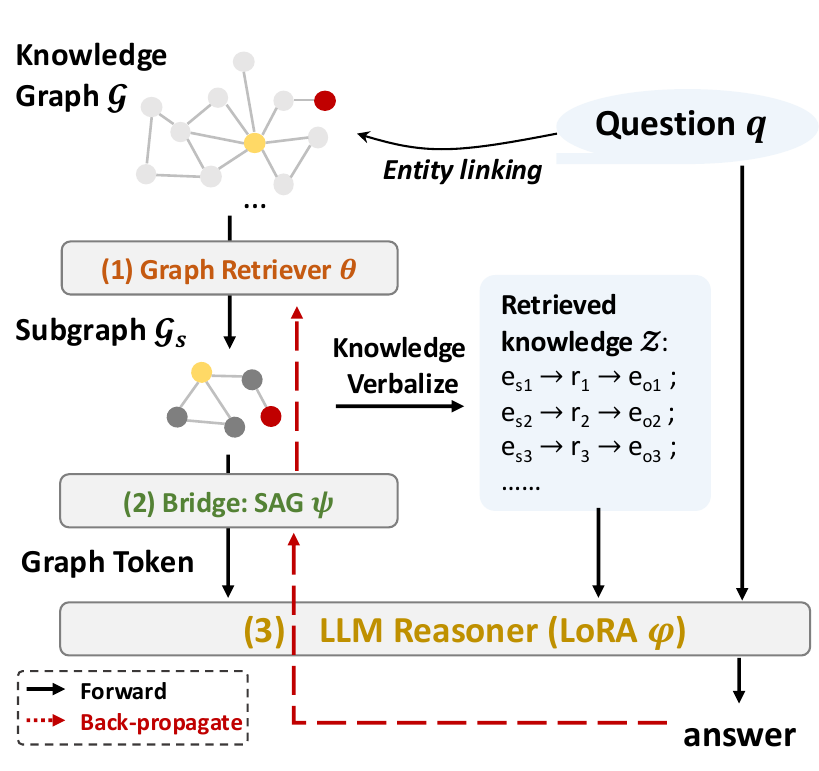}
    \caption{The framework of GRIL.}
    \label{fig:overall}
\end{figure}
 % Starting with a given question $q$ and surrouding knowledge graph $\mathcal{G}$, the graph retriever first retrieves the most relevant subgraph $\mathcal{G}_s$ that is useful for downstream LLM reasoning. Both semantic and structural knowledge within $\mathcal{G}_s$ is infused into LLM through the prompt and soft tokens, respectively. The GNN encoder serves as a bridge between the graph retriever and LLM reasoner, enabling jointly training and interaction between these two modules. entity linking techniques~\cite{dubey2018earl, al2020named}
The overall framework of GRIL is shown in Figure~\ref{fig:overall}. In KGQA tasks, a given question $q$ is typically associated with query entities (\textit{aka} seed entities) in the knowledge graph $\mathcal{G}$. These seed entities are either provided as part of the dataset annotations or could be identified by standard linking procedures~\cite{neumann2019scispacy} when not explicitly available. A graph retriever based on the attention-based growing and pruning mechanism is used to extend the seed entities to a succinct multi-hop subgraph $\mathcal{G}_s\subseteq \mathcal{G}$ (Sec.~\ref{sec:attention_retriever}.) A bridge module encodes the retrieved subgraph through (i) a soft graph token for necessary structural supervision and (ii) verbalized triples for semantic alignment with the LLM (Sec.~\ref{sec:structral_semantic}). These components enable seamless integration with the LLM reasoner, where the LLM’s output logits serve as implicit feedback to train the retriever, enabling end-to-end optimization that encourages the retriever to select the most useful subgraphs for answering the question (Sec.~\ref{sec:joint_training}).

\subsection{Attention-based Graph Retriever}\label{sec:attention_retriever}
\begin{figure}
    \centering
    \includegraphics[width=0.9\linewidth]{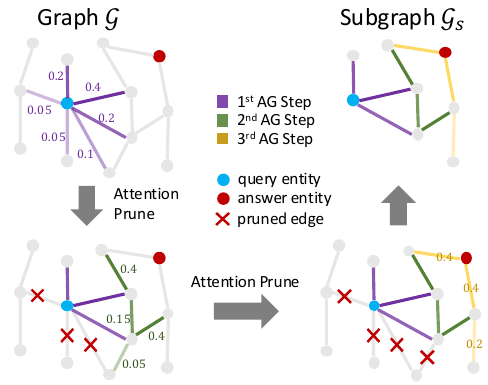}
    \caption{Illustration of Attention-based Graph Retriever. Different colors indicate different steps of Attention Growing (AG Step). Numbers represent the attention scores across neighboring edges. Edges with low attention scores are pruned for model efficiency and better retrieval quality.}
    \label{fig:ar}
\end{figure}

\xhdr{Growing and Pruning Steps}
The knowledge graph can be represented as a set of $N$ fact triplets $\mathcal{G} = \{(e^{(i)}_s, r_{st}^{(i)}, e^{(i)}_t)\}_{i=1}^N$, where $e^{(i)}_s$ and $e^{(i)}_t$ denote the source and target entities, and $r_{st}^{(i)}$ represents the relation, respectively. The attention-based graph retriever dynamically constructs the most relevant knowledge subgraph by employing a growing and pruning mechanism guided by attention scores between entities, as shown in Figure~\ref{fig:ar}. Starting with an initial set of seed entities $E_0$ derived from the question $q$, it computes attention scores $\alpha_{ij}$ between each entity $e_i\in E_0$ and its neighbors $e_j\in\mathcal{N}(E_0)$, where the attention score is calculated by
\begin{equation}\label{eq:attn}
    \alpha_{ij}=\frac{\text{exp}(\text{score}(c_{ij}))}{\sum_{k\in\mathcal{N}(E_0)}\text{exp}(\text{score}(c_{ik}))}, 
\end{equation}
with $\text{score}(c_{ij})=\text{Linear}([h_{e_i}, h_{e_j}, h_{r_{ij}}, h_q])$, and $h_{x}$ indicates the initial representation of $x$ (including entities, relations and the given question) generated by language models, such as SentenceBERT~\cite{reimers2019sentence}. In the growing step, each relation $r_{ik}$ between $v_k\in\mathcal{N}(E_0)$ and $v_i\in E_0$ is associated with an attention score $\alpha_{ik}$, indicating the probability that this relation would be grown from the seed entity $v_i$. Entities in the neighbor set $\mathcal{N}(E_0)$ are added to the entity set $E_1=E_0\cup\{v_k| \alpha_{ik}>0; v_k\in\mathcal{N}(E_0); v_i\in E_0\}$ for the next growing step. In the pruning step, entities with probability scores higher than a certain threshold $\sigma$ are retained in the subgraph, while others are pruned out. The retained nodes will grow to their neighbors in the next growing step. The growing and pruning steps are iteratively performed, ensuring a balance between approaching multi-hop relations and filtering out irrelevant noise. 

\xhdr{Updating Entity Embedding} After each growing and pruning iteration, the graph retriever updates the entity embedding through a message-passing mechanism to aggregate neighbor information.
\begin{equation}\label{eq:entity_update}
h_{e_i}^\prime=W_1h_{e_i} + W_2\sum_{j\in\mathcal{N}(v_i)}\alpha_{ji}h_{e_j},\vspace{-0.3cm}
\end{equation}
where $W_1$ and $W_2$ are learnable weights and $h_{e_i}$ denotes the embeddings of entity $e_i$. $\alpha_{ji}$ is the attention score calculated in the growing step (Eq.~\ref{eq:attn}). The updated embeddings $h_{e_i}^\prime$ are subsequently used for recalculating attention scores in the next iteration. The Entity Embedding Updating step ensures that more relevant neighbors contribute more significantly to the updated entity embeddings, and refines the entity embeddings by incorporating multi-hop contextual information from its neighborhood, which is crucial for accurately modeling complex relationships in the graph.

\xhdr{Subgraph Output} Let $\mathbf{P}\in[0,1]^{|\mathcal{G}|}$ indicate the output probability scores across all relations (\ie edges) in $\mathcal{G}$. The final subgraph $\mathcal{G}_s$ is generated by $\mathcal{G}_s=\mathcal{G}\odot\mathbf{M}$ where the mask matrix $\mathbf{M}$ is sampled conditioned on the probability scores through a differentiable reparameterization trick~\cite{jang2016categorical} as follows,
\begin{equation}
    \mathbf{M}_i=\sigma\left((\log\frac{\epsilon}{1-\epsilon}+\log \frac{\mathbf{P}_i}{1-\mathbf{P}_i})/\tau\right)
\end{equation}
for the $i$-th relation, where $\epsilon\sim \text{Uniform}(0,1)$, $\tau$ is the temperature and $\sigma$ is the sigmoid function. Note that we use the approximately binary matrix $\mathbf{M}$ to achieve the growing and pruning operations during training to make it differentiable, enhancing both model efficiency and training stability.
The detailed algorithm is given in Alg.~\ref{alg:attention} in Appendix~\ref{appd:method}.

\xhdr{Complexity Assessment Module} Since questions across datasets vary in difficulty, they require different amounts of knowledge triplets for reasoning. We thereby propose a Complexity Assessment Module (CAM) that leverages an MLP to predict question complexity, measured by the number of reasoning hops required to reach answer entities. This module takes query embeddings as input and dynamically determines the number of knowledge triplets to provide to the LLM based on the predicted complexity level. The module could be pre-trained and treated as a preprocessing step, offering advantages over using a fixed hyperparameter to specify triplet count, as it automatically adjusts to varying question complexities. We refer to Appendix~\ref{appd:complexity} for more details. 

% $\alpha_{ij}=\frac{\text{exp}(\text{MLP}(h_i^{(L)}))}{\sum_{e_j\in\mathcal{G}_s}\text{exp}(\text{MLP}(h_j^{(L)}))}$, 

\subsection{Semantic and Structural Graph Encoding}\label{sec:structral_semantic}
The bridge module connects the retriever and reasoner by encoding both semantic and structural information, each addressing a distinct and necessary challenge in retrieval-augmented reasoning~\cite{SST}. To encode structural bias, we employ a self-attention graph pooling (SAG)~\cite{SAG} to generate a dense graph-level embedding from the extracted knowledge subgraph $\mathcal{G}_s$. The SAG layer computes self-attention scores $A^{s}\in\mathbb{R}^{|\mathcal{G}_s|\times 1}$, with $A^s_i$ indicating the importance of entity $e_i\in\mathcal{G}_s$, and generates the graph token $h_{\text{GT}}=\operatorname{MLP}(\sum_{e_i\in\mathcal{G}_s} A^s_i h_{e_i}^\prime$) that aggregates global information, where $h_{e_i}^\prime$ denotes the retriever’s contextualized embedding of $e_i$, and an $\operatorname{MLP}$ module projects the contextual embeddings into the same embedding space as the LLM, ensuring dimensional consistency. To incorporate semantic information and facilitate alignment with the LLM's language-based reasoning capabilities~\cite{mavromatis2024gnnrag}, we construct a verbalized subgraph by converting each retrieved triple $(e_s^{(i)}, r_{st}^{(i)}, e_t^{(i)})$ into a natural language format, expressed as $<e^{(i)}_s \rightarrow r_{st}^{(i)}\rightarrow e^{(i)}_t>$. These verbalized triples are concatenated with the original question as follows.
\begin{mdframed}[linewidth=0.5pt]
\texttt{[Graph Token] Based on the following reasoning paths, please answer the given question. \textbackslash n Reasoning
Paths: $e^{(1)}_s \rightarrow r_{st}^{(1)}\rightarrow e^{(1)}_t;\cdots; e^{(n)}_s \rightarrow r_{st}^{(n)}\rightarrow e^{(n)}_t $ \textbackslash n Question: \{Question\} \textbackslash n Answer: \{Answer\} }
\end{mdframed} 
\texttt{[Graph Token]} indicates the soft token derived by the SAG bridge from the global structure within the extract knowledge subgraph. \{Question\} and \{Answer\} are replaced by the question and answer in a certain sample, respectively. In implementation, let $h_{\text{IS}}$ represent the token embeddings of the verbalized triplets and the question. We prepend the soft graph token to the embeddings of the input sequence to form $[h_{\text{GT}}||h_{\text{IS}})]$ as the final LLM input. The soft token is essential for end-to-end training, as it enables the reasoner to influence subgraph selection by optimizing the retriever through task-driven feedback. Together, the soft token and verbalized subgraph allow the reasoner to leverage both structural and semantic information, facilitating coherent reasoning in complex question-answering tasks.

% This graph token is appended as a soft token to the LLM input, which is essential for end-to-end training, as it enables the reasoner to influence subgraph selection by optimizing the retriever through task-driven feedback. 

% Given the retrieved subgraph $\mathcal{G}_s = \{(e^{(i)}_s, r_{st}^{(i)}, e^{(i)}_t)\}_{i=1}^K$, we follow the previous works~\cite{ROG,mavromatis2024gnnrag} and verbalize the subgraph in a list of triplets to add semantics, where each triplet is represented as $<e^{(i)}_s \rightarrow r_{st}^{(i)}\rightarrow e^{(i)}_t>$ in natural language. 
\subsection{Joint Training of Retriever and Reasoner}\label{sec:joint_training}
% \xhdr{General Training Objective}
% The overall input for the LLM reasoner consists of the soft token for graph structural encoding, the verbalized subgraph for semantics, and the original question. This comprehensive input allows the reasoner to leverage both structural and semantic information for accurate predictions. 
Extracting a sub-graph from a massive KG is a discrete, non-differentiable operation. Previous \textit{GNN-as-Retriever} approaches~\cite{mavromatis2022rearev, mavromatis2024gnnrag, jiang2022unikgqa} circumvented this by ranking candidate sub-graphs with a frozen GNN and later fine-tuning an LLM on question-answer pairs. These methods typically rely on answer entities as the training labels, which requires additional cost for the entity labeling and focus merely on relevance regardless of whether the retrieved entities are truly useful for LLM reasoning.  Moreover, it leads to a limitation in open-domain settings where answers are often free-form text rather than explicit KG entities. Instead, we propose to involve implicit feedback from LLMs as a supervision signal that not only optimizes the relevance of the retrieved knowledge but also ensures its maximal utility for the reasoning needs of the LLM. The training loss for this joint system is defined as \vspace {-0.3cm}
\begin{align}
\mathcal{L}_\text{joint} &= \max_{\phi, \psi} \log P_{\phi, \psi}(a | \mathcal{G}_s, q) \label{eq:1} \\
& + \max_{\theta} \log(P_{\phi, \psi}(a | \mathcal{G}_s, q) P_{\theta}(\mathcal{G}_s|q)) \label{eq:2}\vspace{-0.5cm}
\end{align}
$\phi$ and $\psi$ are associated with the LLM reasoner and the SAG bridge, while $\theta$ pertains to the graph retriever. By Bayes' rule, the second part (Eq.~\ref{eq:2}) is equivalent to maximizing the posterior $p_{\phi, \psi, \theta}(\mathcal{G}_s|q,a)$, which involves the additional information from the ground truth answer $a$ compared with the traditional objective Eq.~\ref{eq:log_prob}. $\log(P_{\phi, \psi}(a | \mathcal{G}_s, q))$ is estimated by the LLM reasoner, reflecting its feedback on the quality of the retrieved subgraph $\mathcal{G}_s$. We apply a stop-gradient operator to stop updating the LLM reasoner and the SAG bridge when computing $\log(P_{\phi, \psi}(a | \mathcal{G}_s, q))$, ensuring that the gradient flows correctly during back-propagation. Therefore, the retriever and reasoner are enriched with the inductive bias from the retrieved knowledge subgraph. The LLM ($\phi$) is finetuned with LoRA~\cite{hu2021lora} conditioned on the retrieved subgraph $\mathcal{G}_s$ and question $q$.
% This loss function ensures that the retriever is designed to prioritize relevant subgraphs that not only match the textual query but also complement the logical reasoning required for answering the question. 

\xhdr{Graph Supervision}
In scenarios where answer entities are present in the given KG, prior works~\cite{mavromatis2022rearev,yasunaga2021qa} utilize answer entities as positive labels to supervise the graph retriever. In contrast, our approach expands the set of positive labels to include entities along the shortest paths between the query and answer entities. Specifically, for each question-answer pair ($q, a$), we extract $\mathcal{P}(q,a)$ as the set of entities that lie on any shortest path between the query entities ($q$) and gold answer entities ($a$) in the KG and supervise the retrieved subgraph $\mathcal{G}_s$ to cover $\mathcal{P}(q,a)$ with a binary cross-entropy loss. This additional graph supervision loss is crucial for guiding the retrieval process, as it helps establish the logical connections for effective reasoning. GRIL is trained with both graph supervision loss and the joint loss $\mathcal{L}_{\text{joint}}$. This dual supervision strategy ensures that retrieval prioritizes not just structural relevance but also logical coherence. In open-domain scenarios where traditional methods fail due to the absence of explicit answer entities, GRIL is trained with a single $\mathcal{L}_{\text{joint}}$ containing LLM feedback, which serves as an alternative supervision signal and enables generalizable retrieval in weakly supervised scenarios.

%% file: chapters/04experiments.tex
\section{Experiments}
\subsection{Datasets }
We evaluate GRIL on KGQA tasks. Given a question $q$, the task is to extract relevant subgraphs from the given KG and leverage them for reasoning to get the answer $a$. We conduct experiments on WebQuestionsSP (WebQSP)~\cite{webqsp} and Complex WebQuestions (CWQ)~\cite{CWQ}. WebQSP contains 4,737 natural language questions and CWQ contains 34,699 total complex questions. Both are answerable with a subset Freebase KG within up to 2-hop for WebQSP and up to 4-hop for CWQ. Moreover, we test in the open-domain scenario, where answers might not be explicitly present in the KG. For this more challenging setting, we use the MedQA dataset to evaluate the proposed framework. MedQA contains 12,723 medical questions about disease diagnosis. MedQA is accompanied by the USMLE database and 18 medical textbooks. We manually curate a knowledge graph from the medical textbooks. More details are given in Appendix~\ref{appd:dataset}.

% \subsection{Implementation and Setup}
% We use the Amazon Bedrock to finetune the LLM backbones. All the experiments are implemented with PyTorch on NVIDIA RTX A100 40GB GPUs. Standard dataset splits are applied to each dataset. Following previous studies, we use Hits@1 and F1 as the evaluation metrics. Hits@1 measures the proportion of questions whose top-1 predicted answer is correct. F1 instead balances the precision and recall of the predicted answers\footnote{All the datasets and LLMs used in the submission are public and have been pre-approved.}.

\subsection{Implementation}
 All the experiments are implemented with PyTorch on NVIDIA RTX A100 40GB GPUs. Standard dataset splits are applied to each dataset. Detailed hyperparameter setting is given in Appendix~\ref{appd:hyper_setting}. Following previous studies, we use Hits@1 and F1 as the evaluation metrics. Hits@1 measures the proportion of questions whose top-1 predicted answer is correct. F1 instead balances the precision and recall of the predicted answers. Experimental results in this paper are averaged from three runs with different random seeds.
 % \footnote{We are working with the legal team to release the code upon paper acceptance.}.
\subsection{Experimental Results}
\begin{table}[]\vspace{-0.3cm}
\centering 
\caption{Performance comparison of different methods on the two KGQA benchmarks. We compare with \textit{LLM-only}, \textit{GNN-only}, LLM-as-Retriever (\textit{L-as-R}) and GNN-as-Retriever (\textit{G-as-R}) baselines.}\label{tab:exp1}\vspace{-0.3cm}
\resizebox{\linewidth}{!}{
\begin{NiceTabular}{llcccc} \toprule
              & & \multicolumn{2}{c}{WebQSP}    & \multicolumn{2}{c}{CWQ}       \\
                & & Hits@1           & F1            & Hits@1           & F1            \\ \midrule
\multirow{4}{*}{\rotatebox[origin=c]{90}{\textit{GNN-only}}}&GraftNet & 66.4& 60.4& 36.8& 32.7\\
& SR+NSM+E2E &69.5& 64.1& 49.3 &46.3\\
& UniKGQA        & 77.2          & 72.2          & 51.2          & 49.1          \\
&ReaRev&76.4 & 70.9 & 52.9 & 47.8 \\
  \hdashline
\multirow{4}{*}{\rotatebox[origin=c]{90}{\textit{LLM-only}}}&Llama2-7B      & 64.4          & -             & 34.6          & -             \\
&Llama3-8B      & 65.2          & -             & 35.8          & -             \\
&ChatGPT        & 66.8          & -             & 39.9          &  -             \\
&ChatGPT+CoT    & 75.6          & -            & 48.9          &  -             \\ 
\hdashline
\multirow{5}{*}{\rotatebox[origin=c]{90}{\textit{L-as-R}}}&KD-CoT         & 68.6          & 52.5          & 55.7          & -     \\
&ToG+Llama2-70B & 68.9          & -             & 57.6          &  -             \\
&ToG+ChatGPT    & 76.2          & -             & 58.9          &  -             \\
&RoG            & 85.7          & 70.8          & 62.6          & 56.2          \\
&ToG+GPT4       & 82.6          & -             & \textbf{69.5}          &  -             \\
&EffiQA+GPT4    &82.9           & -             & \textbf{69.5}       & - \\
 \hdashline
\multirow{3}{*}{\rotatebox[origin=c]{90}{\textit{G-as-R}}}&G-Retriever &70.1&-&-&-\\
& GRAG&72.7 &-&-&- \\
& GNN-RAG        & \underline{85.7}          & 71.3          & 66.8          & 59.4          \\
\rowcolor{Gray}
&\textbf{GRIL (8B)}           & \textbf{86.8} & \textbf{73.0} & \underline{68.3} & \textbf{60.5} \\ \bottomrule
\end{NiceTabular}}\vspace{-0.5cm}
\end{table}
\xhdr{Baselines} The baselines can be categorized into four types: (1) \textit{GNN-only} methods, including GraftNet~\cite{graftnet}, NSM~\cite{NSM}, UniKGQA~\cite{jiang2022unikgqa} and ReaRev~\cite{mavromatis2022rearev} that solely rely on graph neural networks for reasoning; (2) \textit{LLM-only} approaches including Llama~\cite{llama3} and ChatGPT~\cite{achiam2023gpt} that utilize LLMs without graph structure; where Llama is fine-tuned on training set.(3) \textit{LLM-as-Retriever} methods including KD-CoT~\cite{KDCOT}, ToG~\cite{TOG}, EffiQA~\cite{dong2024effiqa}, RoG~\cite{ROG} that leverage LLMs to generate relevant relation paths; and (4) \textit{GNN-as-Retriever} approaches~\cite{hu2024grag,he2024g,mavromatis2024gnnrag} that employ GNNs for retrieval and LLMs for reasoning. We select Llama3-8B as the LLM reasoner, while the proposed graph retriever is agnostic to any LLM reasoners.

\xhdr{Results} The results on WebQSP and CWQ are shown in Table~\ref{tab:exp1}. Compared to prior \textit{LLM-as-Retriever} methods, GRIL demonstrates significant improvements in both retrieval accuracy and reasoning quality. Importantly, GRIL achieves comparable or even better performance than leading pipelines (\eg EffiQA and ToG) that rely on proprietary LLMs like GPT-4, which raise potential concerns around accessibility and limited adaptability for domain-specific customization. In contrast, GRIL is built entirely on a smaller, open-source 8B LLM, yet still achieves superior results, showcasing the power of integrating structured retrieval with LLM reasoning in an end-to-end framework. Moreover, GRIL consistently outperforms \textit{GNN-as-Retriever} methods by incorporating LLM feedback directly into retriever training, with a $1.35\%$ average improvement over previous baseline~\cite{mavromatis2024gnnrag}. The joint optimization allows GRIL to identify more relevant subgraphs and retrieve information better aligned with the LLM’s reasoning trajectory, resulting in more precise and robust QA performance across both datasets.
% \begin{table}[ht]\vspace{-0.3cm}
% \centering
% \caption{Performance comparison between different LLM reasoners and Retrievers on WebQSP dataset}\label{tab:LRGR}
% \resizebox{0.8\linewidth}{!}{
% \begin{NiceTabular}{l|l|cc} \toprule
% LLM Reasoner& Retriever& \multicolumn{1}{c}{Hits@1}   & \multicolumn{1}{c}{F1} \\ \midrule
% \multirow{5}{*}{Mistral-7B} & None& 61.3& -\\
% & ES& 76.8& 45.2\\
% & RoG& 83.6&70.0\\
% & GNN-RAG&83.4&70.4\\
% & \textbf{GRIL}$_{\scriptsize \text{seperate}}$&84.3& 71.1\\
% & \textbf{GRIL}$_{\scriptsize \text{end-to-end}}$& \textbf{85.2}& \textbf{70.6}\\
% \midrule
% \multirow{5}{*}{Llama-8B}& None& 65.2&-\\
% & ES& 75.7& 41.8\\
% & RoG&86.3&71.9\\
% & GNN-RAG&85.4&71.2\\
% & \textbf{GRIL}$_{\scriptsize \text{seperate}}$& 86.4& 72.3\\
% & \textbf{GRIL}$_{\scriptsize \text{end-to-end}}$& \textbf{86.8}& \textbf{73.0}\\
% \bottomrule
% \end{NiceTabular}}\vspace{-0.4cm}
% \end{table}

\begin{table}[ht]
\centering
\caption{Hits@1 performance of different retrievers with two LLM reasoners on WebQSP dataset}\label{tab:LRGR}
\resizebox{0.75\linewidth}{!}{
\begin{NiceTabular}{l|cc} \toprule
Retriever & Mistral-7B & Llama-8B \\ \midrule
None & 61.3 & 65.2 \\
ES & 76.8 & 75.7 \\
RoG & 83.6 & 86.3 \\
GNN-RAG & 83.4 & 85.4 \\
\rowcolor{Gray}
\textbf{GRIL}$_{\scriptsize \text{separate}}$ & 84.3 & 86.4 \\
\rowcolor{Gray}
\textbf{GRIL}$_{\scriptsize \text{end-to-end}}$ & \textbf{85.2} & \textbf{86.8} \\
\bottomrule
\end{NiceTabular}}
\end{table}

Table~\ref{tab:LRGR} presents Hits@1 performance comparing different retriever methods paired with different LLM reasoners on the WebQSP dataset. We implement Embedding Similarity (ES), which employs dot-product similarity on contextualized representations from RoBERTa-large~\cite{liu2019roberta} as a dense retrieval approach. All LLM reasoners are fine-tuned with the respective retrieval to ensure a fair comparison. Across both Mistral-7B and Llama3-8B, GRIL demonstrates superior performance, outperforming other retrievers including ES, RoG (\textit{LLM-as-retriever}), and GNN-RAG (\textit{GNN-as-retriever}), which highlights the robustness and generalizability of the proposed method.

\xhdr{Why End-to-End Wins}  GRIL$_{\scriptsize \text{separate}}$, which removes the graph soft token and thereby disables LLM feedback (Eq.~\ref{eq:2}), shows a noticeable performance drop. This suggests that retrieval quality is significantly enhanced through joint optimization with the LLM. Overall, the end-to-end training within GRIL offers two key advantages: (1) it eliminates the need for separately training the retriever, reducing compute costs; and (2) it generalizes better to settings where ground-truth entities are unavailable, as the retriever can be learned directly through supervision from LLMs' signals. These benefits make GRIL particularly useful for open-domain and weakly supervised scenarios.

% As a result, GRIL’s performance drops across datasets. This degradation underscores the critical role of LLM feedback and joint training in guiding the retriever to identify more useful entities and paths, thereby enhancing the overall reasoning capability and retrieval precision. 

\subsection{Open-domain Scenario: MedQA}
\xhdr{Baselines}
We compare GRIL against a diverse set of strong baselines spanning classical information retrieval, dense retrieval, and hybrid methods. We compare with Embedding Similarity (ES) based on dot-product similarity. BM25~\cite{robertson2009probabilistic} leverages exact lexical matching through TF-IDF statistics, which remains competitive in many retrieval scenarios. Contriever~\cite{izacard2021unsupervised} employs contrastive learning with large-scale pretraining on web documents, while SPECTER~\cite{cohan2020specter} leverages scientific documents for more accurate representation. BMRetriever~\cite{xu2024bmretriever} is specifically designed for biomedical information retrieval and pre-trained on massive medical data. All baselines are implemented using the official codebases to ensure fair comparison.
\begin{table}[ht]
\centering
\caption{Accuracy ($\%$) on MedQA\label{tab:medqa}. $\dagger$ indicates that the retriever requires pre-training on massive data. } \vspace{-0.2cm}
\resizebox{0.8\linewidth}{!}{
\begin{tabular}{lcc}
\toprule
Model & Llama2-7b     & Llama3-8b     \\ \midrule
\multicolumn{3}{l}{\textbf{\textit{Without Retriever}}} \\
Zero-shot                   & 42.3$\pm$1.8           & 60.8$\pm$1.6         \\
Fine-tuned & 44.8$\pm$1.0   & 61.7$\pm$1.2 \\    
\hdashline 
\multicolumn{3}{l}{\textbf{\textit{With Retriever}}} \\
\quad ES& 47.6$\pm$1.7          & 63.0$\pm$1.5          \\
\quad BM25& 48.6$\pm$1.6          & 62.5$\pm$1.3          \\
\quad  Contriever$^\dagger$& 49.3$\pm$1.4          & 63.1$\pm$1.0          \\
\quad  SPECTER$^\dagger$& 51.8$\pm$0.9          & 66.2$\pm$1.1          \\
\quad BMRetriever$^\dagger$  &   57.4$\pm$1.7       &  68.9$\pm$1.1    \\
\rowcolor{Gray}
\quad \textbf{GRIL} & \textbf{58.9}$\pm$1.3 &  \textbf{70.4}$\pm$1.6 \\ \bottomrule
\end{tabular}}
\end{table}

\xhdr{Results} The results in Table~\ref{tab:medqa} highlight GRIL's outstanding performance on MedQA dataset, where most baselines from Table~\ref{tab:exp1} fail due to their reliance on answer entities for retriever training. Our GRIL achieves the highest accuracy across both Llama2-7B and Llama3-8B, significantly outperforming all the baselines, including retrievers that have been massively pre-trained on large-scale data. These results emphasize how GRIL operates effectively in open-domain scenarios, where no predefined answer entities are available, and the retrieval process must dynamically adapt to new and unseen queries. The improvement can be attributed to GRIL's structural awareness achieved by the graph retriever and reverse feedback loop from the LLM, key features that distinguish it from conventional RAG methods.

\section{More Evaluation and Analysis}
\subsection{Ablation Study}
\begin{figure}[h]\vspace{-0.3cm}
    \centering
    \includegraphics[width=0.9\linewidth]{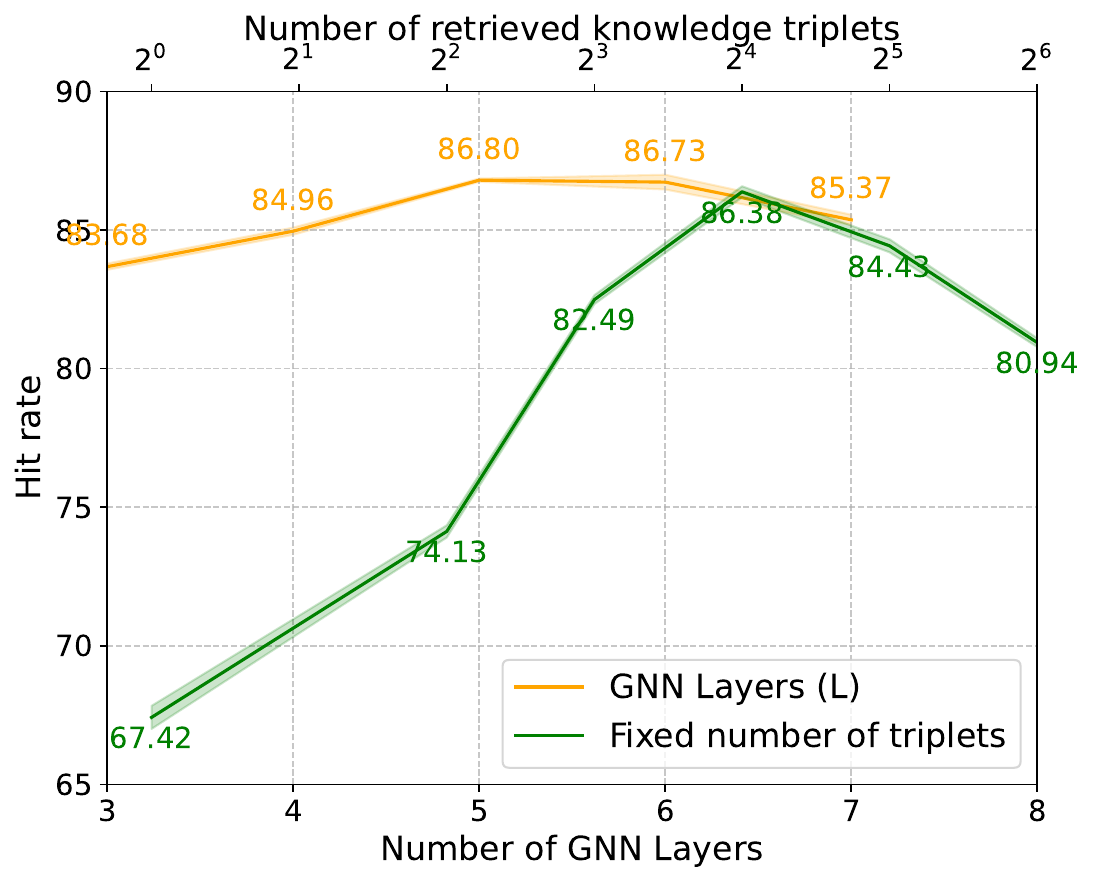}
    \caption{Importance of GNN depth and Complexity Assessment Module on WebQSP dataset}
    \label{fig:hyper-sensitivity}
\end{figure}

\xhdr{GNN Depth Sensitivity and Complexity Assessment Module (CAM) Impact} We conduct a sensitivity analysis on the number of GNN layers to assess the importance of the Complexity Assessment Module (CAM), as shown in Figure~\ref{fig:hyper-sensitivity}. Specifically, we evaluate model performance across varying GNN depths, ensuring the CAM remains consistently integrated (yellow line). Additionally, we compare this to an alternative approach where a fixed number of triplets is retrieved (\eg 16, 32, 64), bypassing the CAM (green line). As shown in the figure, when using a fixed number of triplets, performance initially improves with more triplets but peaks and declines beyond 16 triplets. Instead, CAM successfully addresses the sensitivity with a more stable performance (yellow line). CAM dynamically adjusts the number of retrieved triplets based on the question complexity level, achieving improved performance while avoiding the computational overhead introduced by excessive retrieval.
% , underscoring its effectiveness in tailoring the graph-based reasoning process to varying question complexities.

\begin{figure*}[]
    \centering
    \includegraphics[width=0.95\linewidth]{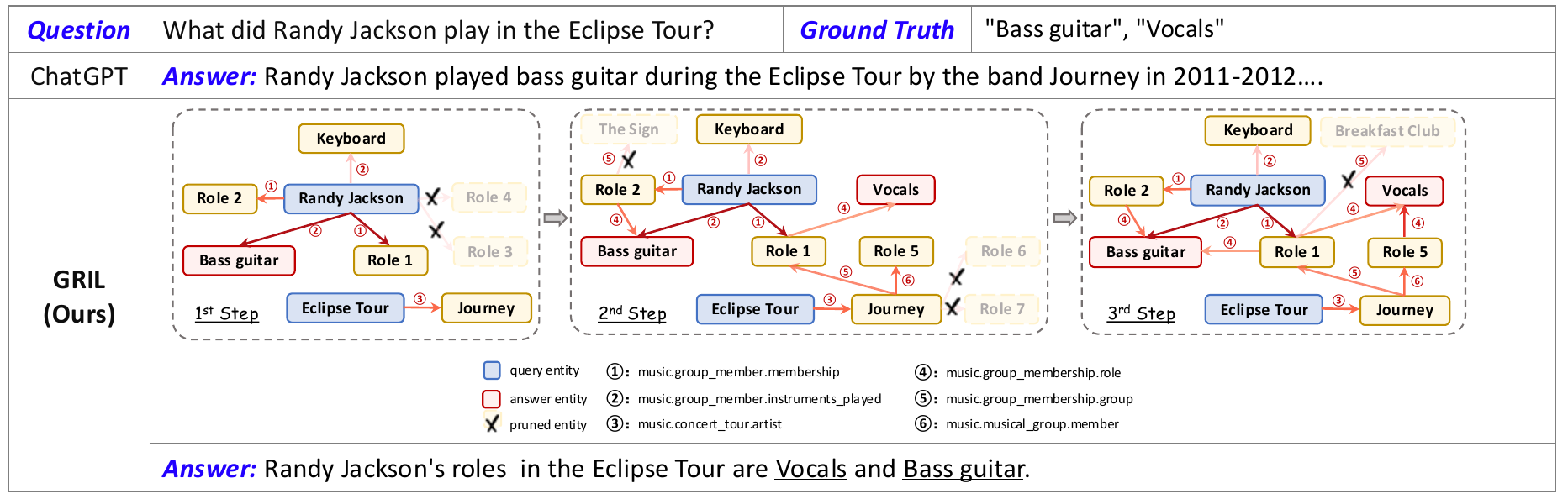}
    \caption{A case study of GRIL retrieval and reasoning on CWQ dataset. In the Retrieved Subgraph, the edge color intensity indicates the importance score of the certain knowledge triplet.}
    \label{fig:case}
    \end{figure*}
\xhdr{Ablation on Graph Retriever} Table~\ref{tab:ablation_attention} presents the effects of different graph pruning mechanisms (\eg threshold-based or top-$K$ strategy) and specific operators on the WebQSP dataset. When using threshold-based pruning, increasing the threshold tends to slightly improve inference efficiency while reducing the F1 scores. $\sigma=0.1$ yields the highest F1 and a balance between performance and efficiency, which is set as default in GRIL.
Removing key mechanisms, such as the pruning operation and the entity embedding updating step (Eq.~\ref{eq:entity_update}) in the graph retriever, results in significant performance drops. The absence of pruning also substantially increases inference time and destroys the performance, underscoring its importance in maintaining retrieval efficiency and precision. 
% as it requires fewer hyperparameters to tune and allows us to flexibly control the pruning process with a uniform parameter $\sigma$. This flexibility avoids the constraints of the top-$K$ strategy, which would need to be adjusted dynamically for each step. 
\begin{table}[h]
\centering
\caption{Ablation study on the graph retriever} \label{tab:ablation_attention}
\resizebox{0.9\linewidth}{!}{
\begin{tabular}{lll} \toprule
&F1 &inference time (\textit{s}) \\
\midrule
\multicolumn{3}{l}{\textbf{\textit{Pruning Mechanism Alternatives}}}\\
Threshold ($\sigma=0.1$)            & \textbf{72.68}         & 0.476                             \\
Threshold ($\sigma=0.2$)             & 71.82$_{(\downarrow1.18\%)}$& 0.463$_{(\downarrow2.73\%)}$                             \\
Threshold ($\sigma=0.5$)             & 71.61$_{(\downarrow1.47\%)}$& 0.423$_{(\downarrow11.13\%)}$\\
Top 5                       & 72.54$_{(\downarrow0.19\%)}$& 0.459$_{(\downarrow3.57\%)}$\\
Top 10                      & 72.60$_{(\downarrow0.11\%)}$& 0.463$_{(\downarrow2.73\%)}$\\
Top 20                      & 72.37$_{(\downarrow0.43\%)}$& 0.479$_{(\uparrow0.63\%)}$\\
\hdashline
\multicolumn{3}{l}{\textbf{\textit{Removing Key Steps in Graph Retriever}}}\\
\textit{w/o} pruning       & 70.28$_{(\downarrow3.30\%)}$& 0.687$_{(\uparrow44.33\%)}$                           \\
\textit{w/o} Entity Update & 70.32$_{(\downarrow3.25\%)}$& 0.437$_{(\downarrow8.19\%)}$\\ 
% \textit{w.o.} LLM Training & 72.30$_{(\downarrow0.52\%)}$& 0.476$_{(-)}$ \\
\bottomrule                
\end{tabular}}
\end{table}

\subsection{Combination with Small LMs}
\begin{table}[ht]
\centering
\caption{Performance comparison between small language models and LLMs as the reasoner.} \label{tab:smallLM}
\resizebox{\linewidth}{!}{
\begin{NiceTabular}{lcccc} \toprule
                          & \multicolumn{3}{c}{\textbf{WebQSP}}  \\ \midrule
                          & Hits@1           & F1            &     Time (\textit{s})& Size \\
BERT-large& 40.8& -& 0.97 &$\sim$ 336M \\
RoBERTa-large& 41.3& -& 0.87 & $\sim$ 355M\\
Llama2-7B& 64.4& -& 4.02 &$\sim$ 7B \\
Llama3-8B& 65.2& -& 3.87 &$\sim$ 8B\\
% \textbf{GRIL}  & \textbf{86.4} & \textbf{72.6} & 4.22 \\
GRIL \textit{w/} BERT     & 64.8          & 70.2          & 1.54 &$\sim$ 768M \\
GRIL \textit{w/} RoBERTa  & 67.7          & 71.4          & 1.32 &$\sim$ 806M \\
\bottomrule                
\end{NiceTabular}}
\end{table}
Table~\ref{tab:smallLM} compares the performance of small language models (LMs) and LLMs as the reasoner on the WebQSP dataset. GRIL, paired with a small LM, \eg RoBERTa or BERT, demonstrates comparable or even better performance to LLMs, while incurring significantly lower inference costs. The observations justify that graph-based reasoning capability compensates for the reduced model size without sacrificing accuracy. This result underscores the parameter efficiency of our approach, as it effectively harnesses the graph retriever to enhance reasoning and retrieval quality, making it highly practical for resource-constrained scenarios. Importantly, GRIL’s retrieval cost grows with the size of the extracted subgraph, not with the full knowledge graph (when the entity embedding updating step is disabled). In addition, the Complexity Assessment Module (CAM) predicts the minimal hop depth required for each query, effectively bounding the retrieval scope. This design keeps inference latency stable even as the full KG size increases, ensuring that GRIL scales efficiently to large graphs.

\subsection{Case Studies}
We conduct a case study on CWQ dataset to illustrate the effectiveness of GRIL on retrieval and complex reasoning, as shown in Figure~\ref{fig:case}. While ChatGPT provides partially correct answers, it fails to capture crucial aspects of the ground truth (\ie ``vocals'' in the case). Instead, GRIL successfully identifies the necessary logical connections between the question entity (``Randy Jackson'') and the answer entity (``Vocals'').
We visualize how GRIL navigates the knowledge graph to establish connections between relevant entities, through iterative growing and pruning. It clearly highlights the predicate relationships that lead to a comprehensive and accurate answer generated by the LLM reasoner. Moreover, the edge color intensity represents the importance score of each edge at its corresponding growing step, further enhancing the GRIL's self-interpretability and providing insights into the underlying reasoning logic of GRIL.

%% file: chapters/05conclusion.tex
\section{Conclusion}
In this work, we present GRIL, a novel framework that integrates graph retrieval and reasoning through attention-based growing and pruning mechanisms and joint training with LLMs. Our approach enhances reasoning over knowledge graphs and eliminates the need for predefined answer entities, making it highly effective in open-domain scenarios. Experimental results show significant performance improvement on KGQA tasks and superior scalability, while also improving inference efficiency. GRIL provides a cost-effective, state-of-the-art solution for knowledge-intensive tasks and offers potential for real-world applications.

\xhdr{Limitations and Future Work}
While the proposed GRIL shows promise, several aspects could be further explored to extend its applicability. First, GRIL assumes that the graph structure is inherently necessary for question answering, as it relies on structured knowledge graphs for multi-hop reasoning. However, this assumption may limit its ability to handle scenarios where the underlying knowledge is either unstructured or where the graph structure does not fully capture the complexity of natural language semantics. Future work could explore ways to automatically and organically integrate \textit{GNN-as-Retriever} and \textit{LLM-as-Retriever} approaches, enabling the model to dynamically determine when to leverage graph-based reasoning and when to rely on unstructured, text-based retrieval. Moreover, extending GRIL to other applications that require graph reasoning, such as recommendation systems and biomedical knowledge extraction, may reveal additional challenges and opportunities for improvement.

%% file: chapters/06appendix.tex
\appendix
\section{Dataset}\label{appd:dataset}
We evaluate GRIL on three question answering datasets: WebQSP~\cite{webqsp}, CWQ~\cite{CWQ} and MedQA~\cite{jin2021disease}. Detailed dataset statistics are shown in Table~\ref{tab:dataset_stat}.

\xhdr{WebQSP and CWQ} Both datasets are designed for question answering tasks that leverage the Freebase knowledge graph~\cite{bollacker2008freebase}, which comprises over 164.6 million facts and 24.9 million entities. WebQSP primarily requires up to 2-hop reasoning to answer questions, whereas CWQ presents a more complex challenge, necessitating up to 4-hop reasoning over the provided knowledge graph. We follow the previous setting~\cite{ROG,he2021improving,jiang2022unikgqa}\footnote{\url{https://huggingface.co/datasets/rmanluo/RoG-webqsp}}\footnote{\url{https://huggingface.co/datasets/rmanluo/RoG-cwq}} for knowledge graph extraction. Specifically, the input knowledge graph for each question is constructed by a subset of Freebase KG that contains all triples within the max reasoning hops of question entities. We follow the previous studies~\cite{ROG} for dataset split. The initial seed entities are derived from the question and provided in the dataset. Both benchmarks are inherently tied to a specific knowledge graph (KG) and our experiments are conducted using the same KG provided by the respective benchmarks. 

\xhdr{MedQA} is a 4-way multiple-choice medical question-answering task, originating from practice tests for the United States Medical License Exams (USMLE), which generally require a deep understanding of related medical concepts from associated medical textbooks.  We utilize the original dataset split setting~\cite{jin2021disease}, with 80\% for training, 10\% for development, and 10\% for test. For MedQA, we use a self-constructed knowledge graph based on the 18 given medical textbooks. Specifically, we use ScispaCy~\cite{neumann2019scispacy} to identify biomedical concepts as entities and the \textit{RecursiveCharacterTextSplitter} from
LangChain\footnote{\url{https://www.langchain.com}} to split the medical textbooks into snippets. Edges are created between two entities that are mentioned within one snippet. The embeddings of entities and relations are initialized using Sentence PubmedBert~\cite{pubmedbert}. Note that we do not consider the effect of different entity recognition tools and splitters, as they are orthogonal to the focus of this work. Experimental comparisons with baselines are based on the same curated knowledge graph.

Compared with the knowledge graph in previous studies~\cite{yasunaga2021qa} built on the Disease
Database of the Unified Medical Language System (UMLS)~\cite{bodenreider2004unified} and DrugBank~\cite{wishart2018drugbank}, our curated knowledge graph demonstrates a significantly improved answer coverage, increasing from $24.6\%$ to $88.4\%$.
\begin{table}[H]
\centering \caption{Dataset statistics. Coverage indicates the }\label{tab:dataset_stat}
\resizebox{\linewidth}{!}{
\begin{NiceTabular}{lcccc} \toprule
Dataset & Train & Dev & Test  & Coverage(\%)\\
\midrule
WebQSP & 2,848 & 250 & 1,639 & 94.9 \\ 
CWQ&27,639&3,519 & 3,531 & 79.3 \\
MedQA& 10,178& 1,272 & 1,273 & 88.4\\
\bottomrule
\end{NiceTabular}}
\end{table}

\section{Training Details}\label{appd:hyper_setting}
We set the hidden size as 512 in the graph retriever and the GNN encoder. The batch size is set to 2 during training and 4 during evaluation. The learning rate is 1e-5. The maximum number of epochs is 100. An early stopping strategy is used to mitigate overfitting. We utilize the LoRA~\cite{hu2021lora} technique to finetune the LLM reasoner with rank $8$ by default. All experiments are conducted with PyTorch on NVIDIA RTX A100 GPUs for three runs with different random seeds.

\section{Additional Module Details}
\subsection{Attention-based Graph Retriever}\label{appd:method}
The detailed algorithm of our attention-based graph retriever is shown in Alg.~\ref{alg:attention}. The attention scores are calculated for each edge (\ie knowledge triplet). To avoid useless attention growing for irrelevant entities and keep the focus on important entities, the graph retriever iteratively performs growing and pruning steps. After the attention calculation, if the number of triplets with attention scores lower than $\sigma=0.1$ is larger than a certain budget (\eg 16), then the retriever automatically performs the pruning step. The hyperparameter sensitivity of $\sigma$ is shown in Table~\ref{tab:ablation_attention}. Moreover, the entity embeddings are updated by message-passing and aggregating mechanisms with the previously calculated attention scores. The output of the algorithm contains $G$ with updated entity embeddings and probabilities on each triplet $\mathbf{P}$. The final subgraph $\mathcal{G}_s$ is generated by $\mathcal{G}_s=\mathcal{G}\odot\mathbf{M}$ where the mask matrix $\mathbf{M}$ is sampled conditioned on the probability scores through a differentiable reparameterization trick~\cite{jang2016categorical}.
\begin{algorithm*}[t]
\caption{Attention-based Graph Retriever}\label{alg:attention}
\SetAlgoLined
\KwIn{Knowledge Graph $G = \{ (e^{(i)}_s, e^{(i)}_t, r_{st}^{(i)}) \}_{i=1}^N$, Query $q$, Seed Entity $e_0$, Number of Layers $L$}
\KwOut{Probability on triplets, Updated $G$}
\BlankLine
Initialize zero $\mathbf{E} \in \mathbb{R}^N$  \tcp*{Initialize probability on triplets}
$P_1 \gets \text{get-neighbor}(\{ e_0 \})$ \tcp*{Retrieve initial list of neighbor triplets}
\For{$i = 1, 2, \dots, L$}{
    $\mathbf{A}_i \gets \text{AttnScore}(P_i, q_0, G) \in [0, 1]^{|P_i|}$ \tcp*{Compute Attention Scores (Eq.~\ref{eq:attn}) on $P_i$}
    \If{Pruning step}{
        $\mathbf{A}_i \gets [\mathbf{A}_i > \sigma]$ \tcp*{Keep attention scores greater than threshold $\sigma$}
    }
    $P_{i+1} \gets \text{get-neighbor}(P_i, \mathbf{A}_i)$ \tcp*{Update neighbors based on attention weights}
    $\mathbf{E} \gets \text{update}(\mathbf{E}, \mathbf{A}_i) \in \mathbb{R}^N$ \tcp*{Update probability on triplets}
    \For{$v \in P_i$}{
        $\mathcal{N}(v) \gets \{\text{neighbors of } v \text{ in } P_i \text{ with non-zero attention}\}$
        
        $\mathbf{m}_v \gets \sum_{u \in \mathcal{N}(v)} \mathbf{A}_i[u] \cdot \text{Message}(e_u,r_{uv})$ \tcp*{Aggregate messages}
        
        $e_v \gets \text{Update}(e_v, \mathbf{m}_v)$ \tcp*{Update entity embedding}
    }
}
\Return $\mathbf{E}$, $G$ with updated embedding
\end{algorithm*}

\subsection{Complexity assessment module}\label{appd:complexity}
We approach the complexity assessment as a classification task, utilizing a multilayer perceptron (MLP) to predict the question complexity (i.e., the number of reasoning hops) based on the query embedding generated by language models. The ground truth is defined as the shortest path distance between the query entities and the answer entities. The MLP is trained using cross-entropy loss, comparing the predicted number of hops with the ground truth. Table~\ref{tab:complexity_predictor} presents the prediction accuracy (\%) when the model is trained on individual datasets or a combined dataset of WebQSP and CWQ. Notably, BERT outperforms RoBERTa in all settings, achieving higher accuracy on both individual datasets (WebQSP and CWQ) as well as the joint dataset (WebQSP+CWQ). The joint dataset consistently yields the best results, with BERT achieving the highest accuracy of 74.28\%, showcasing the benefits of combining diverse reasoning tasks to improve generalization. We select BERT as the language model and train the MLP on the combined dataset as the default Complexity Assessment Module. Given the predicted number of hops, $c$, for a specific question, we allocate $5 \times c$ as the number of final retrieved triplets to be provided for downstream reasoning. Notably, this module can be pre-trained and treated as a preprocessing step, enhancing its efficiency. Alternatively, a fixed hyperparameter can be employed to specify the number of retrieved triplets, offering a trade-off for reduced computational overhead.
\begin{table}[H]
\centering \caption{Accuracy ($\%$) of number of hops prediction}\label{tab:complexity_predictor}
\resizebox{\linewidth}{!}{
\begin{NiceTabular}{lccc} \toprule
 Dataset & WebQSP & CWQ   & WebQSP+CWQ \\ \midrule
RoBERTa & 63.33  & 70.28 & \textbf{73.46}      \\
BERT  & 64.98  & 72.36 & \textbf{74.28}     \\
\bottomrule
\end{NiceTabular}}
\end{table}

\subsection{Retrieval Augmentation Ensemble}
Retrieval augmentation (RA)~\cite{mavromatis2024gnnrag} enhances the performance of LLM reasoners by aggregating knowledge retrieved through different mechanisms. Building on the previous work~\cite{mavromatis2024gnnrag}, we extend the \textit{GNN-as-Retriever} paradigm by incorporating an \textit{LLM-as-Retriever} approach to further enrich the retrieval process. Specifically, we integrate reasoning paths retrieved from RoG~\cite{ROG}, complementing them with those retrieved by our graph-based method. This union of reasoning paths combines the strengths of both graph-based and language-based retrieval, thereby expanding the diversity of knowledge incorporated into the reasoning process. As a result, this approach not only broadens the scope of relevant information but also enhances the robustness and accuracy of the overall LLM reasoning mechanism. Beam-search decoding is used in \textit{LLM-as-Retriever} approaches to generate diverse reasoning paths for better answer coverage. We set the number of beams as 3 in RoG and report the performance of Retrieval augmentation (RA) in Table~\ref{tab:RA}.
\begin{table}[H]
\centering \caption{KGQA Performance with and without RA on WebQSP and CWQ dataset}\label{tab:RA}
\resizebox{\linewidth}{!}{
\begin{NiceTabular}{lcccc} \toprule
& \multicolumn{2}{c}{WebQSP}    & \multicolumn{2}{c}{CWQ}       \\
& Hits@1& F1& Hits@1& F1\\ \midrule
GNN-RAG & 85.7&71.3&66.8&59.4 \\
GNN-RAG+RA & 90.7&\textbf{73.5}&68.7&60.4\\
GRIL&86.8&73.0&68.3&60.5 \\
GRIL+RA&\textbf{91.4}&73.1&\textbf{69.2}&\textbf{61.8} \\
\bottomrule
\end{NiceTabular}}
\end{table}
We observe that Retrieval augmentation (RA) consistently improves the KGQA performance, with an average improvement rate of $3.58\%$ on WebQSP and $1.99\%$ on CWQ.  GRIL demonstrates superiority when paired with RA. For example, GRIL+RA achieves the highest Hits@1 on both WebQSP and CWQ, outperforming GNN-RAG+RA by significant margins of $0.7\%$ and $0.5\%$, respectively. While GRIL without RA already outperforms baselines on both datasets, RA further enhances its performance. This demonstrates GRIL’s ability to better exploit the additional reasoning paths provided by RA, particularly in the more complex CWQ dataset, which features longer and more intricate question-answering dependencies.

\subsection{Ablation on Textual Subgraph}
We further study the role of textualizing the retrieved subgraph. In standard retrieval-augmented frameworks, converting the retrieved knowledge into natural language is a widely adopted practice, since LLMs are inherently trained to process text rather than raw graph embeddings. To quantify its effect, we compare GRIL with and without the textual representation of subgraphs on MedQA dataset.
\begin{table}[h]
\centering
\caption{Ablation study on the textualization of the retrieved subgraph.}\label{tab:ablation_TS}
\resizebox{\linewidth}{!}{
\begin{tabular}{lcc}
\toprule
\textbf{Model} & \textbf{Llama2-7B} & \textbf{Llama3-8B} \\
\midrule
GRIL  & \textbf{58.9} & \textbf{70.4} \\
\qquad \textit{w/o} soft token    & 53.6 & 66.1 \\
\qquad \textit{w/o} textual subgraph& 50.7 & 64.8 \\
\bottomrule
\end{tabular}}
\end{table}
From Table~\ref{tab:ablation_TS}, we observe that removing the textual representation and the graph soft token both lead to a significant performance drop, highlighting their importance and necessity in transferring semantic and structural information.